\documentclass[letterpaper, 10 pt, conference]{ieeeconf}  

\IEEEoverridecommandlockouts                            

\usepackage{amsmath} 
\usepackage{amssymb}  
\usepackage{mathtools}
\usepackage{isomath}
\usepackage{multirow}
\usepackage{url}
\usepackage{graphicx}
\usepackage{color}
\usepackage[font={normal}]{caption}

\usepackage{cite}
\usepackage{float}
\usepackage{algorithm}
\usepackage{algpseudocode}
\usepackage{setspace}
\usepackage{dblfloatfix}

\usepackage{multirow}
\usepackage{lipsum}
\usepackage{tabularx}
\newcolumntype{P}[1]{>{\centering\arraybackslash}p{#1}}
\newcolumntype{M}[1]{>{\centering\arraybackslash}m{#1}}
\newcolumntype{C}[1]{>{\centering\arraybackslash}m{#1}}

\algrenewcommand\algorithmicindent{0.6em}

\algnewcommand{\CustomComment}[1]{ \hspace*{37.5mm}{#1}}

\DeclareCaptionFont{mysize}{\fontsize{8}{9.6}\selectfont}
\usepackage{footnote}

\title{\LARGE \bf
Pushing in the Dark: A Reactive Pushing Strategy for Mobile Robots Using Tactile Feedback
}

\author{ Idil Ozdamar$^{*1,2}$, Doganay Sirintuna$^{*1,2}$, Robin Arbaud$^{1,2}$, and Arash Ajoudani$^1$
\thanks{$^*$ Contributed equally to this work.}
\thanks{$^1$HRI$^2$ Lab, Istituto Italiano di Tecnologia, Genoa, Italy. {\tt\small idil.ozdamar@iit.it}, {$^2$ Dept. of Informatics, Bioengineering, Robotics, and System Engineering. University of Genoa, Genoa, Italy.}}
\thanks{This work was supported by the European Research Council's (ERC) starting grant Ergo-Lean (GA 850932).}
}

\begin{document}

\maketitle
\thispagestyle{empty}
\pagestyle{empty}

\begin{abstract}
For mobile robots, navigating cluttered or dynamic environments often necessitates non-prehensile manipulation, particularly when faced with objects that are too large, irregular, or fragile to grasp. The unpredictable behavior and varying physical properties of these objects significantly complicate manipulation tasks. To address this challenge, this manuscript proposes a novel \textit{Reactive Pushing Strategy}. This strategy allows a mobile robot to dynamically adjust its base movements in real-time to achieve successful pushing maneuvers towards a target location.
Notably, our strategy adapts the robot motion based on changes in contact location obtained through the tactile sensor covering the base, avoiding dependence on object-related assumptions and its modeled behavior.
The effectiveness of the \textit{Reactive Pushing Strategy} was initially evaluated in the simulation environment, where it significantly outperformed the compared baseline approaches. 
Following this, we validated the proposed strategy through real-world experiments, demonstrating the robot capability to push objects to the target points located in the entire vicinity of the robot.
In both simulation and real-world experiments, the object-specific properties (shape, mass, friction, inertia) were altered along with the changes in target locations to assess the robustness of the proposed method comprehensively.  

\end{abstract}

\section{Introduction}\label{sec:introduction}

Continuous developments in mobile manipulator technology have reached a point where their utilization is integral across a broad spectrum of application areas, including factories, warehouses, and space exploration. To execute effective operations in these domains, it is necessary for robots not to be constrained by limited manipulation capabilities. Therefore, pushing becomes a crucial motion primitive that expands robot manipulation abilities \cite{survey}, especially for tasks like clearing paths \cite{taskin} or transporting objects that are large, unwieldy, or ungraspable \cite{unwieldy}, particularly when robot lacks an arm, has limited payload capacity, or is already holding an object \cite{Ellis}.

\begin{figure}[!t]
\centering
\includegraphics[width=0.75\columnwidth]{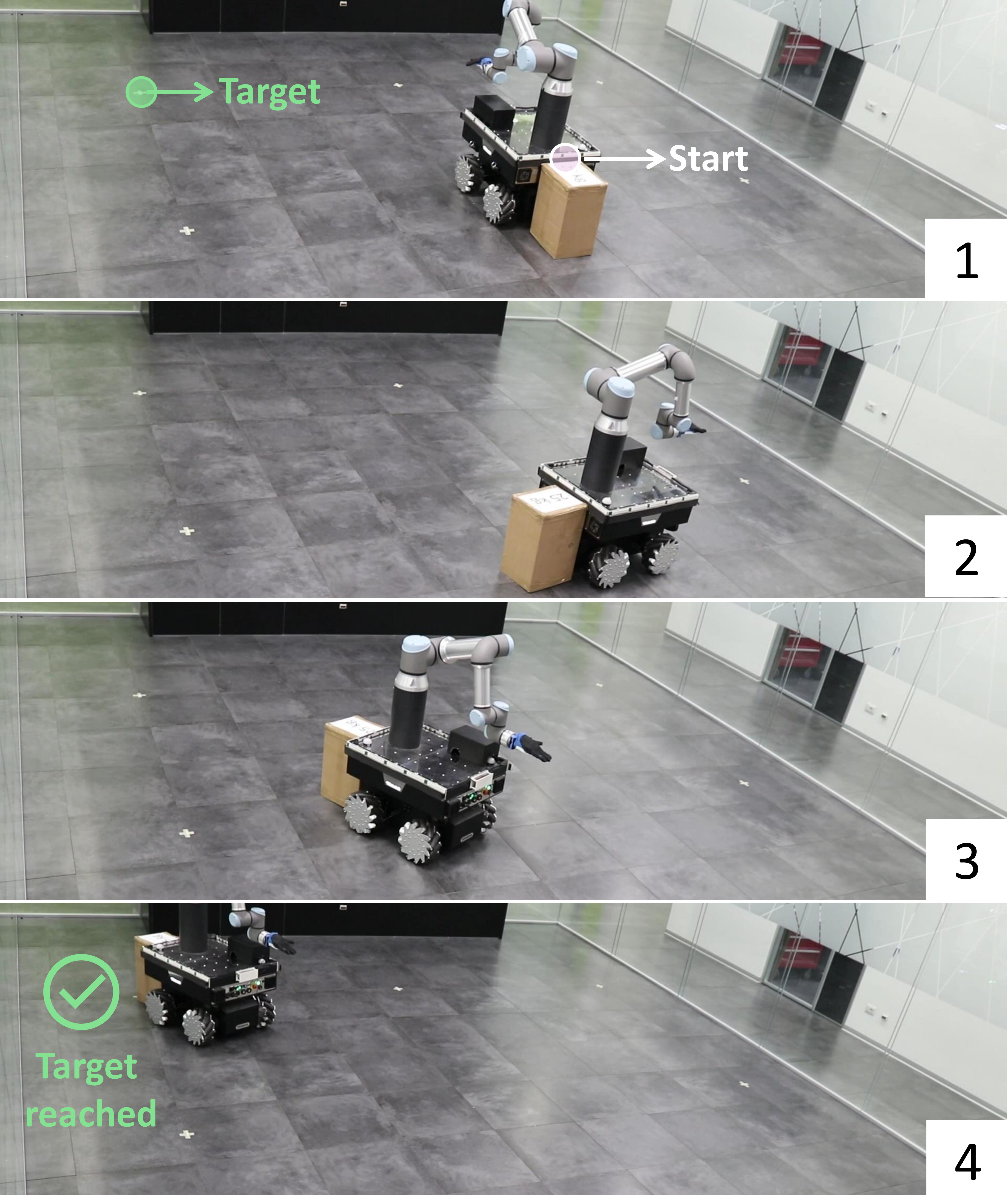}\vspace{-2mm}
\caption{Snapshots demonstrating the robot pushing a heavy box (25 kg) toward a target location positioned behind it.}
\label{fig:intro}
\vspace{-8mm}
\end{figure}

Although nonprehensile manipulation and pushing with legs are intuitive tasks for humans, despite the diverse and sometimes anisotropic properties of the objects, this skill remains a challenge for robots \cite{LetsPushReview}. Therefore, much of the state-of-the-art depends on apriori knowledge or simplifications that may not always accurately represent the behavior of most objects in real-world pushing scenarios. Following the introduction of the limit surface concept \cite{Goyal1991} that shows the relationship between the motion of the pushed object and the friction forces,  Lynch and Mason \cite{Lynch_Mason} developed an open-loop planner explicitly relying on frictional assumptions. Here, the planner aims to maintain sticking line contact between the pusher and the slider, called stable pushing, using the known friction value and assumption of uniform friction properties over the support plane. Similarly, an open-loop approach that depends on the knowledge of the object model was proposed in \cite{Zhou2019PushingRevisited}, where sticking single-point contact was preserved throughout the pushing action. 

Since the relative robot-object position does not change when the sticking constraint is imposed by using the friction cone assumptions, repositioning actions would be eliminated during pushing. Therefore, nonprehensile transportation studies utilizing non-holonomic bases often incorporate stable pushing. For instance, Bertoncelli et al. \cite{Bertoncelli2020} integrated pushing stability constraints into their linear time-varying MPC scheme to maintain a stiff contact. In a more recent study \cite{unwieldy}, a simplified version of this controller was introduced to address the computationally expensive nature of the problem, offering an alternative approach for utilizing an open-source solver for nonlinear model predictive control (NMPC). In addition, the authors proved that establishing a line contact, as opposed to a point contact, is a necessity for nonholonomic robots to attain stable pushing. On the other hand, Hogan et al. \cite{Hogan2020ReactivePN,HoganICRA} demonstrated the effectiveness of relaxing the sticking constraint if the robot is not confined to nonholonomic locomotion. In these studies, where a 6-DoF robotic arm was utilized, an MPC approach in tandem with integer programming was employed to govern both the sticking and slipping behaviors of the object during pushing. Another study \cite{Lynch1992touch}, which similarly does not impose constraints on sticking behavior, proposed a control system based on only tactile sensing to rotate the object to a desired orientation, excluding control over its position.

In contrast to prior studies that presume knowledge of object-specific properties, a line of research has emerged concentrating on learning the behavior of objects when they are pushed \cite{LetsPushReview}. Lau et al. collected hundreds of samples on how the object moves based on the robot’s relative position and pushing direction to build corresponding mappings \cite{Lau2011}. Then, using this mapping, the desired pushing of irregular-shaped objects was accomplished through a non-parametric regression method. In \cite{Ruiz-Ugalde2010, Ruiz-Ugalde2011}, a similar data collection procedure was executed to obtain a mathematical model for planar sliding motion where the static and kinetic friction coefficients between the robot, the support plane, and the object are determined. Alternatively, Hermans et al. \cite{Hermans} focused on estimating the effective contact location to be able to push the unknown objects in a straight line based on a learning method trained with six objects. Another data-driven approach for planar pushing utilized the past pushing experiences in a rapidly exploring random tree (RRT) based planner \cite{Mericli2013, Mericli2015}.
Furthermore, Agrawal et al. \cite{Agrawal} generated a model for pushing that solely depends on raw images to enable the robot to learn through independent exploration without the need for human intervention.

However, to the best of the authors' knowledge, only a few studies tackle the challenge of pushing objects to specific destinations without prior knowledge of the object motion dynamics.
For example, Igarashi et al. \cite{Igarashi2010} introduced a controller merging orbiting around the object and pushing movements, as an analogy to the dipole field in physics, by relying on the vision-based measurements of the object and the robot. 
Building on this concept, Krivic et al. introduced an adaptive pushing skill in \cite{Krivic2016} and expanded it in \cite{Pushing_corridor}, which learns the inverse model of the unknown object interaction on the fly. The experiments with different objects, where a circular holonomic base robot was employed, demonstrated the effectiveness of the vision-based model, which regulates the orbiting and pushing actions.

In this study, we present a novel \textit{Reactive Pushing Strategy} that enables the robot to drive the object towards the goal location through dynamic adjustments in the robot motion according to the changes in the contact location without depending on any other sources of information. 
The main contributions of this work can be summarized as follows:

\begin{itemize}
    \item The introduced strategy can handle objects with varying shapes, masses, and friction properties, even if the contact type (point or line) changes during pushing. As our strategy operates reactively, with the objective of pushing objects without relying on assumptions, prior experience, or a training phase, it presents a versatile and efficient approach for dynamic object transportation in real-world applications.
    \item The developed method eliminates the necessity for object pose information, typically obtained from vision-based measurements, wherein the challenges related to occlusion, tracking accuracy, and other factors might impede the performance. Instead, a tactile sensor,  namely the \textit{Capacitive Touch Sensor} (explained in detail in Section \ref{sec:touch_skin_sensor}), covering the sides of the robot base (see Fig. \ref{fig:framework}), is employed to acquire contact location information of the pushed object. With this, we demonstrated that successful positional control of objects through pushing is possible without requiring additional sensors for object orientation.   
    \item The effectiveness of the \textit{Reactive Pushing Strategy} was assessed in simulation by transporting objects with different shapes, frictions, and inertial properties to target points distributed across the entire area surrounding the robot. Moreover, we compared our strategy with two baseline approaches, one being a non-reactive pushing strategy and the other being an existing method in the literature \cite{Pushing_corridor}.
    \item Lastly, the algorithm was validated through real-world experiments involving challenging goal locations and objects with different physical properties.
\end{itemize}

\begin{figure}[b] \vspace{-5mm}	
        \centering
	\includegraphics[width=0.88\columnwidth]{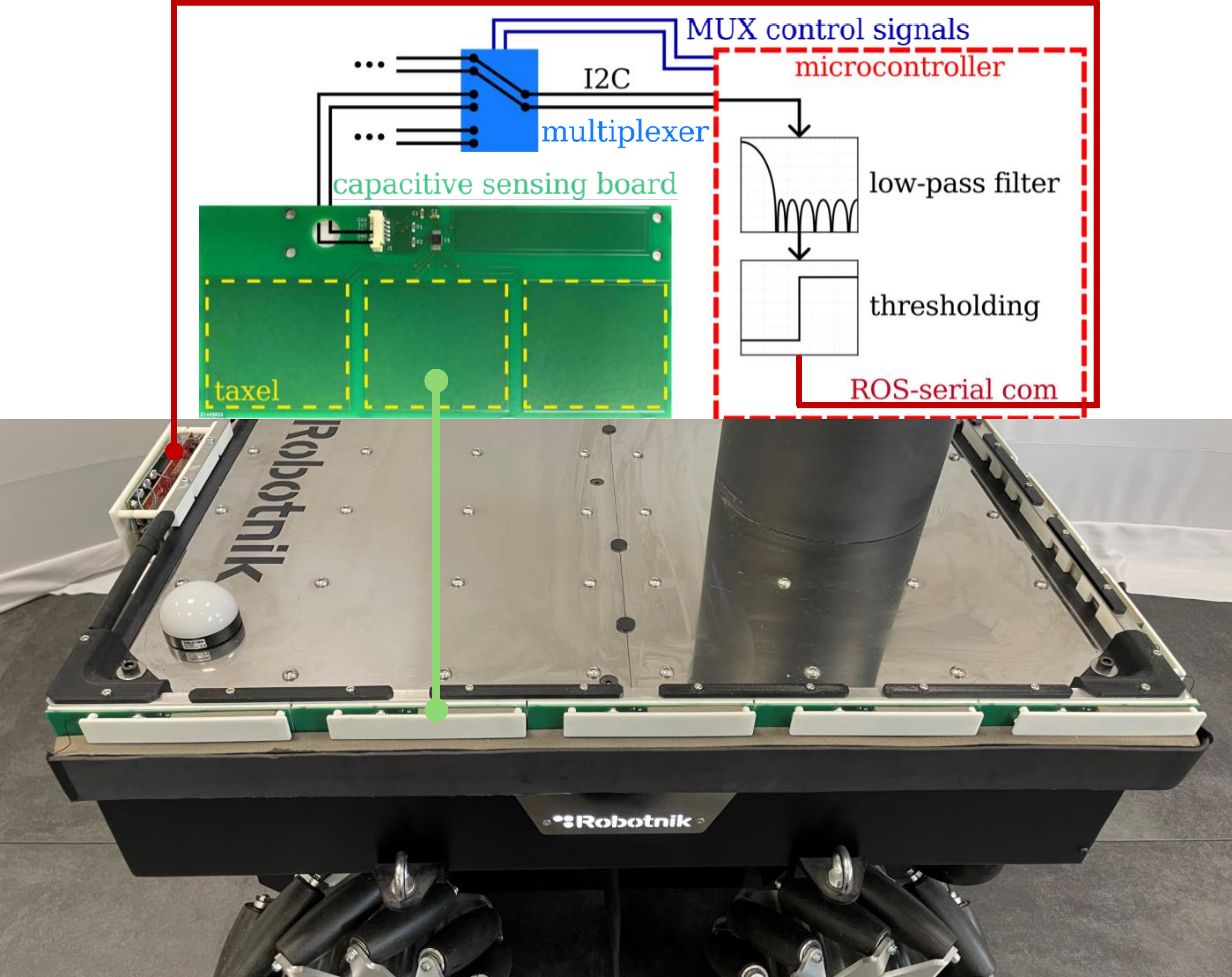}\vspace{-0.2cm}
	\caption{Picture and electrical schematic of the capacitive sensing system.}
	\label{fig:sensor}
\end{figure}

\begin{figure*}[!t]	
	\includegraphics[width=0.99\textwidth]{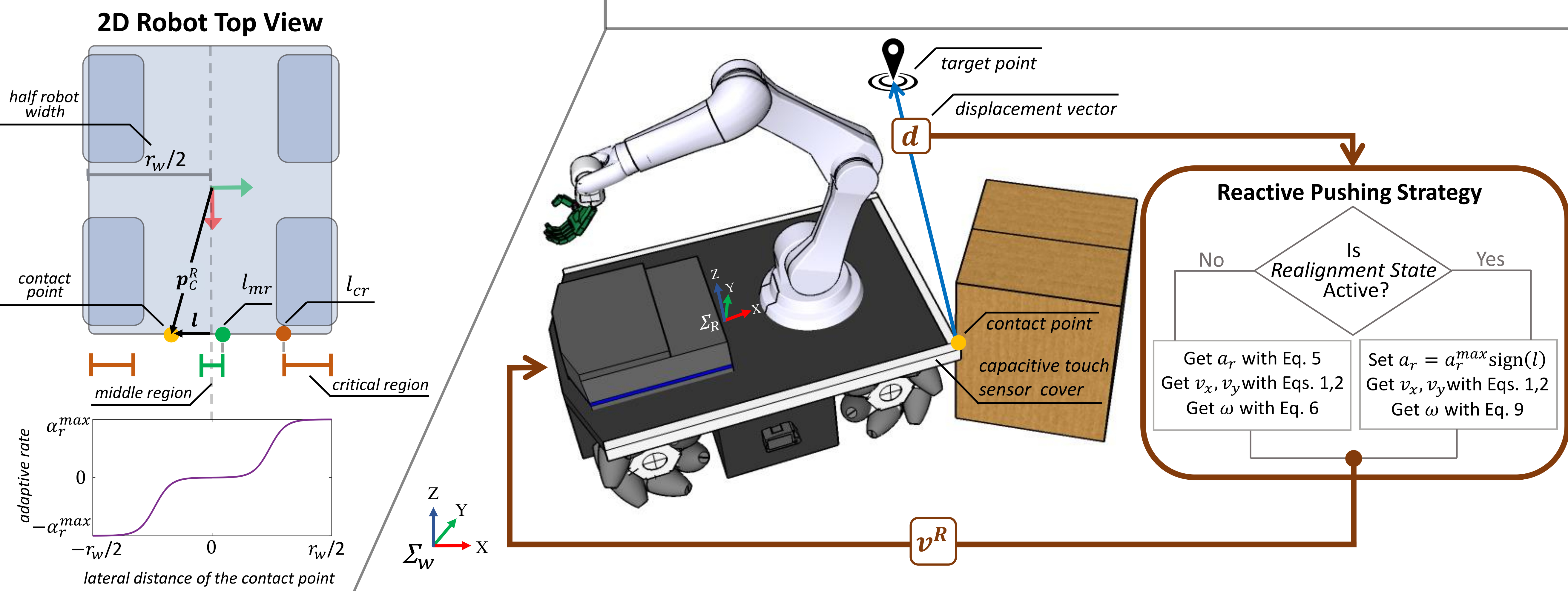}\vspace{-2mm}
	\caption{High-level scheme of the proposed framework.}
	\label{fig:framework}
  \vspace{-0.6cm}
\end{figure*}

\section{Reactive Pushing System Overview }\label{sec:reactive_pushing_system_overview}

\subsection{The Robot Platform}\label{sec:robot_platform}

In this study, the Kairos mobile manipulator serves as the robotic platform to push unwieldy objects towards predefined targets. It consists of a velocity-controlled Robotnik SUMMIT-XL STEEL mobile base and a high payload 6-DoFs Universal Robot UR16e arm with a Pisa/IIT Softhand gripper. Noteworthy, despite the presence of the arm, only the base is employed for the task. Thanks to the omnidirectional wheels of the platform, non-holonomic constraints are eliminated, thereby enabling movements over a large workspace. In addition, the rectangular-shaped base, with a 250 kg payload capacity, makes it suitable for pushing bulky objects.

\subsection{Capacitive Touch Sensor}\label{sec:touch_skin_sensor}

The sensory system shown in Fig. \ref{fig:sensor} is an improved version of the one presented in \cite{MOCA-S}. It is constituted of a frame holding 14 sensing PCBs, and a central system based on the EK-TM4C1294XL development board from Texas Instruments, complemented by an I2C multiplexing board. The sensing PCBs are covered, on the front, left, and right sides of the robot base, with 3 stripes of polyurethane foam coated with a grounded layer of conductive ink. The ink formulation and coating method are the same as in \cite{ecotac}. On each sensing PCB, a chip measures the capacitance between three distinct electrodes and the conductive ink layer.  Thus, 42 sensing areas, or \textit{taxels}, are available. When the robot base comes into contact with an object, the foam gets compressed, and the distance between the ink layer and the electrode under the point of contact decreases, thus increasing the measured capacitance on the corresponding taxel.

The capacitive sensing PCBs are based on the FDC1004 capacitance-to-digital converter from Texas Instruments. This chip I2C address is fixed, so connecting all 14 of them required the use of several distinct I2C buses as well as multiplexing. 5 independent I2C interfaces are used, each of them connected sequentially to 3 different FDC1004 chips through a dual channel multiplexer (SN74CB3Q3253).

The stream of values coming from each taxel is low-pass filtered, then compared to a threshold. When the capacitance exceeds that threshold, the measurement is timestamped and transferred to the robot main computer through ROS-Serial. Following this procedure, the contact location w.r.t the robot frame in X-Y plane
($\boldsymbol{p}^{R}_{C}$ $\in$ $\mathbb{R}^{2}$) is computed based on these values and the respective positions of the taxels. When only one taxel is activated, the contact type is classified as \textit{point}, and $\boldsymbol{p}^{R}_{C}$  is determined as the taxel position. However, if contact is detected by multiple taxels, the contact type is designated as \textit{line}. This form of interaction can be modeled as if only the extreme points are in contact \cite{Xielinecontact}, and the position vector $\boldsymbol{p}^{R}_{C}$ is computed by averaging the positions of the taxels situated at the boundaries of the contact.

\subsection{Reactive Pushing Strategy (RPS)}\label{sec:reactive_pushing_strategy}

The core of our proposed strategy lies in generating reactive movements through touch, without the necessity of prior knowledge regarding the object size, weight, inertia, texture, and consequently its frictional forces. We operate under the assumption of quasi-static pushing movements, which already covers a large proportion of pushing activities in real-world scenarios.

In order to compute the desired velocities of the mobile base, denoted as $\boldsymbol{v}^{R} = [{v}^{R}_x, {v}^{R}_y,\omega]^T$ where ${v}^{R}_x$ and ${v}^{R}_y$  $\in$ $\mathbb{R}$ are the linear velocities in the X and Y directions, and $\omega$ $\in$ $\mathbb{R}$ is the angular velocity, we drew inspiration from the kinematic bicycle model\cite{Corke2017} to avoid abrupt turns that could lead to the loss of contact with the pushed object. Building on this, the \textit{Reactive Pushing Strategy} combines $v^{*}$ $\in$ $\mathbb{R}_{\geq 0 }$ parameter, which regulates the magnitude of the linear velocity, with an adaptive rate, denoted as $a_{r}$ $\in$ $\mathbb{R}$, to generate $v^{R}_{x}$ and $v^{R}_{y}$ given by: \vspace{-2mm}
\begin{equation}\label{eq:v_x}\vspace{-1mm}
    v_{x}^{R} = \frac{v^{*}}{\sqrt{1+a_{r}^2}},
\end{equation}
\begin{equation}\label{eq:v_y}
    v_{y}^{R} = \frac{a_{r}v^{*}}{\sqrt{1+a_{r}^2}},
\end{equation}  
where the value of ${v}^{*}$ is adjusted according to the remaining distance between the contact point and the target. It is formulated as follows:\vspace{-2mm}
\begin{equation}\label{eq:v_star}\vspace{-1mm}
    v^{*} = K_{v}||\boldsymbol{d}||,   
\end{equation}
\begin{equation}\label{eq:distance_vec}
    \boldsymbol{d}  = \boldsymbol{p}^{W}_{T} - (\boldsymbol{p}^{W}_{R} + \boldsymbol{R}^{W}_{R}\boldsymbol{p}^{R}_{C}),
\end{equation}
where $\boldsymbol{d}$ $\in$ $\mathbb{R}^{2}$ is the displacement vector, $K_{v}$ $\in$ $\mathbb{R}_{>0}$ is the velocity gain, $\boldsymbol{p}^{W}_{T}$ and $\boldsymbol{p}^{W}_{R}$ $\in$ $\mathbb{R}^{2}$ represent the target and robot positions w.r.t the world frame, $\boldsymbol{R}^{W}_{R}$ $\in$ $\mathbb{R}^{2\times 2}$ is the rotation matrix between the world and the robot frame.  

Secondly, the adaptive rate is computed on the fly based on the lateral distance between the contact location and the robot center ($l = p^{R}_{C, y}$ $\in$ $\mathbb{R}$) by employing a logistic function. With this function (see the bottom left of Fig. \ref{fig:framework}), minor lateral repositioning motions are generated when the object is near the robot center, and repositioning escalates rapidly as it approaches the corners, as follows:  
\begin{equation} \label{eq:a_r}
    a_{r} = \begin{cases} (\zeta + \frac{(\eta - \zeta)}{1+e^{(\beta-|{l}|)k}})\mathrm{sgn}(l), \hspace{3mm} \text{if} \hspace{3mm} ||\boldsymbol{d}|| > d_{th}
    \\
    0, \hspace{3mm} \text{else}
    \end{cases},
\end{equation}
where $\eta$, $\zeta$, $\beta$, and $k$ $\in$ $\mathbb{R}_{\geq0}$ define the function's maximum and minimum asymptotes, the inflection point, and the steepness of the curve, respectively. Hence, when the contact point is close to the center ($l \approx 0$), the robot generates faster forward motion ($v^{R}_{x}$) to reach the target together while continuing to maintain the object in the middle with lower lateral velocity ($v^{R}_{y}$). On the contrary, as the contact point approaches the edges of the base, the parameter $a_{r}$ enables higher lateral velocity ($v^{R}_{y}$), which facilitates a sliding behavior for the object toward the center of the robot. However, this behavior may not be desirable when the target is in close vicinity since it could inadvertently lead the robot away from the target. For this reason, if the object is within a certain distance threshold ($||\boldsymbol{d}||$ $\leq$ $d_{th}$), we set $a_r$ to zero to cancel the lateral motion, which yields to bring the object much closer to the target point.

In order to drive the robot towards the goal position, the formulation of the desired angular velocity can be given as:
\begin{equation}\label{eq:w}
    \omega = \frac{v_{x}^{R}}{L}\tan{\gamma},
\end{equation}
where $L$ $\in$ $\mathbb{R}_{>0}$ is a scalar term affecting the sharpness of the curvature within the generated motion trajectory and $\gamma$ $\in \mathbb{S}$\footnotemark\footnotetext{Unit circle, set of angles [$-\pi, \pi$) } refers to the steering angle. It is computed as:
\begin{equation}\label{eq:gamma}\vspace{-1mm}
    \gamma = K_{h}(\theta^{*} \ominus \theta),
\end{equation}
\begin{equation}
    \theta^{*} = atan2(d_y, d_x),
\end{equation}
where $K_{h}$ $\in$ $\mathbb{R}_{>0}$ represents the heading gain, with the heading corresponding to the X-axis of the robot, while $\theta$ and $\theta^{*}$ $\in \mathbb{S}$ refer to the robot orientation in the world frame and the heading angle of $\boldsymbol{d}$, respectively. Note that $\ominus$ operator computes the difference between $\theta$ and $\theta^{*}$, representing the heading error, which is constrained within the range of [$-\pi, \pi$).

It is important to note that the change of $\omega$ as a function of $v^{R}_{x}$ in addition to $\gamma$ offers a key advantage to our pushing strategy. When the goal point is distant and the object is close to the center of the robot, the fast forward motion ($v^{R}_{x}$) can allow for high $\omega$ values, resulting in a rapid decrease in the remaining heading error. On the other hand, as the robot approaches the goal, rapid turning movements are avoided due to low $\omega$ since the robot decelerates ($v^{R}_{x}$ gets smaller).

Despite the help of the lateral motions during pushing, if the contact point enters one of the critical regions, which starts after a distance of $l_{cr}$ from the center of the robot (see Fig. \ref{fig:framework}), our proposed strategy activates a particular state, namely the \textit{Realignment State}, to prevent the loss of contact. 
This state remains active until the contact point is successfully re-positioned to the middle region of the robot front side ($l_{mr}$). To accommodate the changes between these states, we defined a dynamic threshold variable ($l_{th}$), whose value switches between $l_{mr}$ and $l_{cr}$.
The linear velocities are computed by utilizing the maximum adaptive rate value ($a_r = a^{max}_r\mathrm{sgn}(l)$) in Eqs. \ref{eq:v_x} and \ref{eq:v_y} to prioritize the lateral movements for swift alignment of the object to the middle region (see Fig. \ref{fig:framework}) of the robot. 

Furthermore, during the ongoing alignment of the object, this state seamlessly integrates the rotational behavior to diminish the heading error. This is accomplished by formulating the $\omega$ as:\vspace{-1mm}
\begin{equation}\label{eq:w_rfm}\vspace{-1mm}
    \omega = \sigma\frac{v^{*}}{L}\tan{\gamma},
\end{equation}
\begin{equation} \label{eq:g_rfm}
  \sigma = \begin{cases}
  1-\dfrac{|l|}{l_{cr}}, \hspace{5mm}  \text{if} \hspace{3mm} 0 \leq |l| \leq l_{cr} \\ 
  0,  \hspace{28mm}  \text{else}
 \end{cases},
\end{equation}
where $\sigma$ $\in$ $\mathbb{R}_{\geq 0 }$ is a gain that is inversely proportional to the deviation from the center. The rationale behind reducing $\omega$ is to mitigate the risk of failing to compensate for the sliding towards the edges due to the high rotational motion, with the lateral velocity ($v^R_y$) that tries to center the object imposed by \textit{Realignment State}. It's important to highlight that in order to prevent the object from escaping the corners, $\omega$ becomes zero (see Eq.~\ref{eq:g_rfm}) when the object is in the critical region ($|l|$ $>$ $l_{cr}$). 

The pseudocode outlined in Algorithm 1 covers the complete sequence of the \textit{Reactive Pushing Strategy}.  It specifies the conditions that govern the state changes and illustrates the associated computations for determining the desired $\boldsymbol{v^{R}}$ sent to the robot base.

\begin{algorithm}[!h]
\caption{Reactive Pushing Strategy}\label{alg:alg_1}
\begin{algorithmic}
\setstretch{1.1}

\Require $\boldsymbol{p}^{W}_{T}$, $\boldsymbol{p}^{W}_{R}$, $\boldsymbol{R}^{W}_{R}$, $\boldsymbol{p}^{R}_{C}$
\Ensure $\boldsymbol{v^{R}}$ 

\State \emph{Initialization:}
\State $l_{th} \gets l_{cr}$
\State \emph{Control loop:}
\State ${l} \gets \boldsymbol{p}^{R}_{C,y}$  

\State ${\boldsymbol{d}} \gets getDisplacementVector$ \Comment{Eq.~\ref{eq:distance_vec}} 

\If{$ ||\boldsymbol{d}|| < {d}_{success}$} \Comment{Target reached}
    \State $\boldsymbol{v^{R}} \gets [0,0,0]$
\Else \Comment{Continue pushing}
    
    \If{$|{l}| > l_{th}$} \Comment{Enter \textit{Realignment State}}
        \State $l_{th} \gets l_{mr}$ 
        \Comment{Update threshold}
        \State $ a_{r} \gets a_r^{max}\mathrm{sgn}(l) $
        \State $ [v_{x}^{R},v_{y}^{R}] \gets getLinearVelocity(v^*, a_{r})$
        \Comment{Eq.~\ref{eq:v_x}\ref{eq:v_y}}
        \State $ w \gets getAngularVelocity(v^*, \sigma, \gamma)$
        \Comment{Eq.~\ref{eq:w_rfm}}

    \Else 
        \State $l_{th} \gets l_{cr}$ 
        \Comment{Update threshold} 
        \State $a_r \gets getAdaptiveRate(l,\boldsymbol{d})$
        \Comment{Eq.~\ref{eq:a_r}}
        
        \State $ [v_{x}^{R},v_{y}^{R}] \gets getLinearVelocity(v^*, a_{r})$
        \Comment{Eq.~\ref{eq:v_x}\ref{eq:v_y}}
     
        \State $ w \gets getAngularVelocity(v_{x}^{R}, \gamma)$
        \Comment{Eq.~\ref{eq:w}}

    \EndIf

\EndIf
\State $\boldsymbol{v^{R}} \gets [v_{x}^{R}, v_{x}^{R},\omega ]$
\Comment{Send velocities to the robot}

\end{algorithmic}
\end{algorithm}
\setlength{\textfloatsep}{16pt}

\section{Experiments and Results}\label{sec:experiments_and_results}

To validate the effectiveness of our proposed system, we assessed the performance of our \emph{Reactive Pushing Strategy} (RPS) through experiments conducted in both virtual and real-world environments. We additionally compare our RPS with a) a non-reactive pushing strategy (NPS), and b) a state-of-the-art method from the literature namely adaptive pushing skill (APS)~\cite{Pushing_corridor}. 

\subsection{The System  Parameters}\label{subsec:simulation_experiments}

For the experiments, the selected parameters of the logistic function (refer to Eq.~\ref{eq:a_r}) that adjusts the adaptive rate ($a_r$) were $\eta = 10$, $\zeta = 0$, $\beta = 0.12$, $k = 60$ (see Fig. \ref{fig:framework} bottom left for the profile). With this formulation, $a_r$ escalates swiftly to 10 when the contact location approaches the edges of the robot front side, resulting in high lateral velocity ($v^{R}_{y}$). Conversely, when the object is in the middle, this value remains approximately 0, facilitating only forward motion to move towards to target. Furthermore, the gains that steer the robot towards to target were tuned based on a heuristic to be $K_h = 1.0$, $K_v = 0.1$, and $L = 0.2$. The value of $l_{cr}$ was set to 24 cm, which established a critical region length of 5 cm near the edges of the robot base. Additionally, a distance of 50 cm was chosen for $d_{th}$. Finally, the velocities computed by our proposed strategy were saturated at 0.05 m/s for linear velocity and 0.15 rad/s for angular velocity prior to being commanded to the robot, aiming to promote a quasi-static pushing motion.

\begin{figure}[!h]\vspace{-2mm}
\centering
\includegraphics[width=0.70\columnwidth, trim=0cm 0cm 0cm 0cm, clip=true]{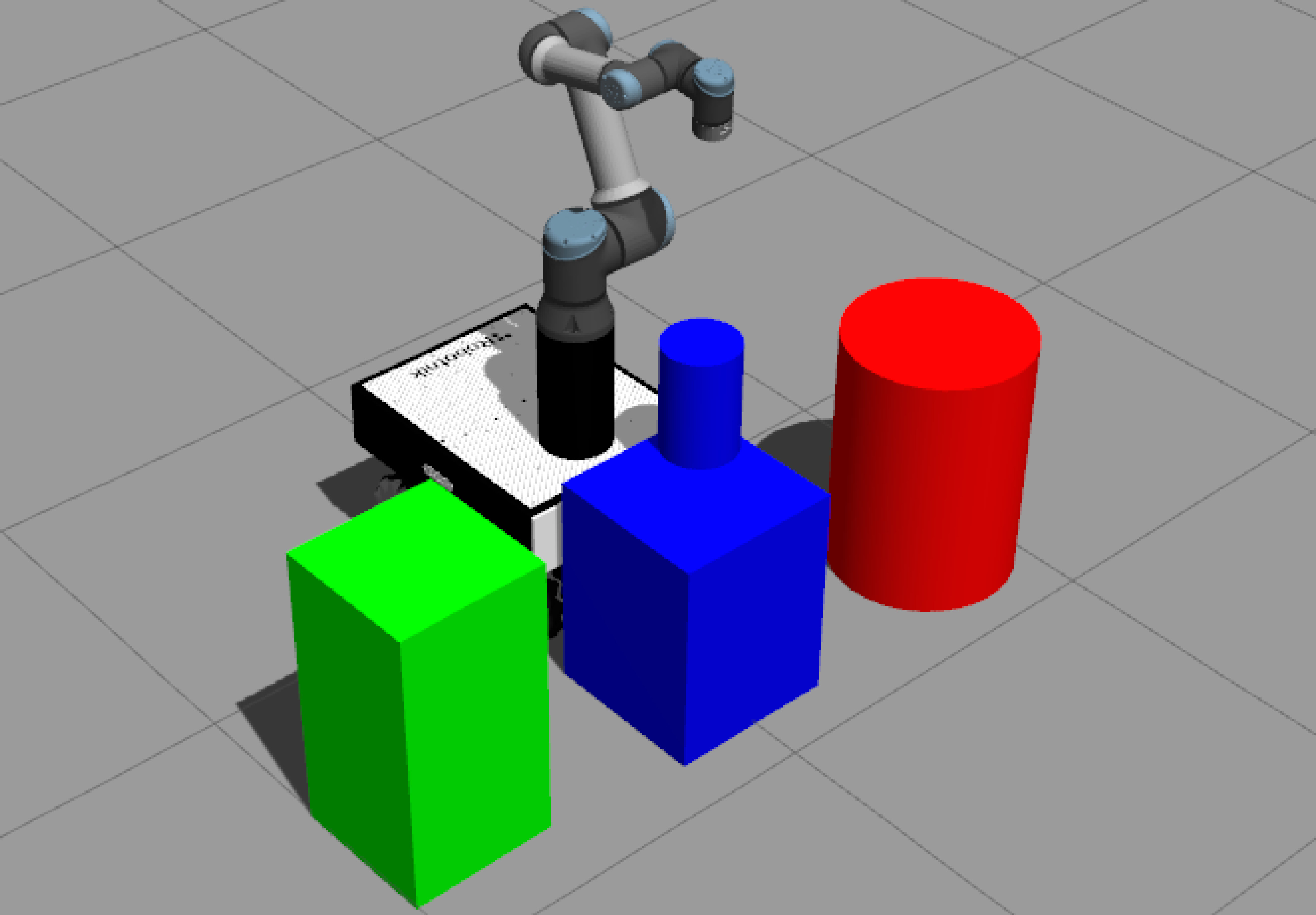}\vspace{-2mm}
\caption{The set of objects pushed in the simulation experiments: a) a 20 kg box with dimensions of 40 $\times$ 40 $\times$ 80 cm, b) a 5 kg box with dimensions of 45 $\times$ 45 $\times$ 60 cm, and a 10 kg cylinder positioned on top of the rear left corner with a radius of 10 cm and height of 30 cm to simulate asymmetric mass distribution, and c) a 25 kg cylinder with a radius of 25 cm and height of 70 cm.}
\label{fig:objects}
\vspace{-0.6cm}

\end{figure}

\begin{figure*}[!t]	
	\includegraphics[width=0.96\textwidth]{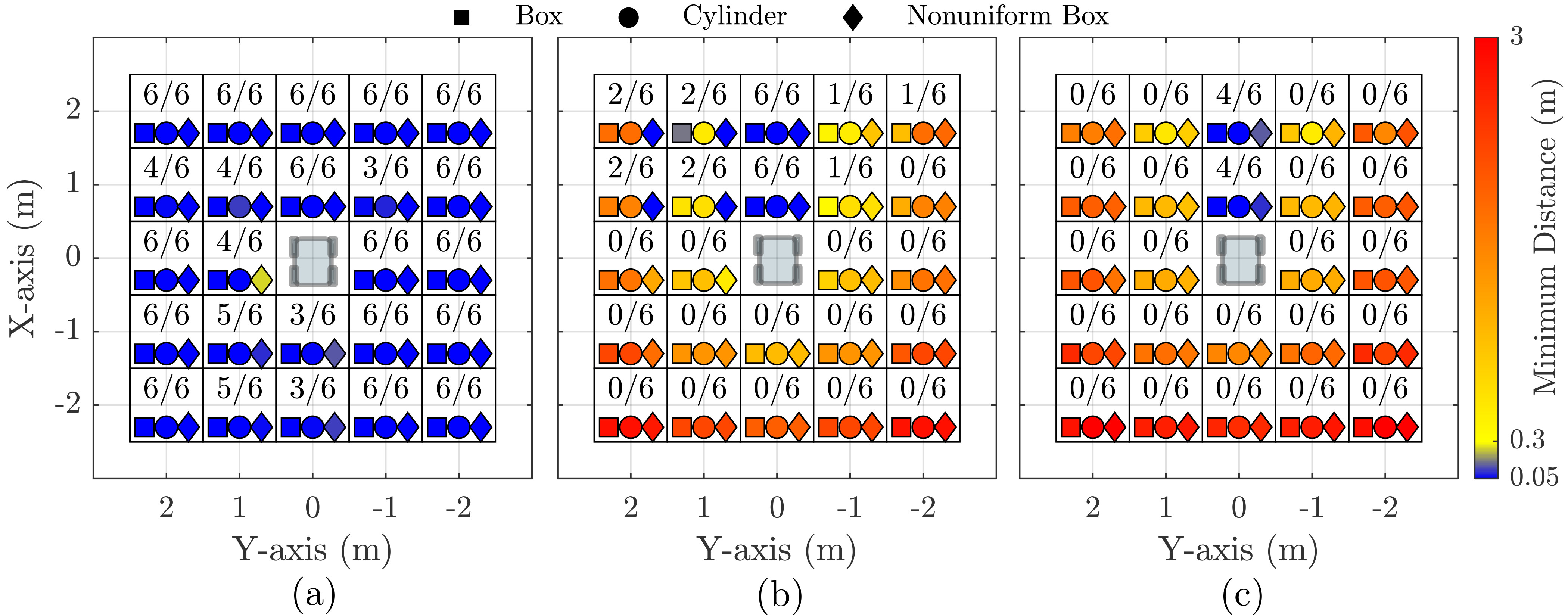}
    \vspace{-0.2cm}
	\caption{Results of the simulation experiments with a) RPS, b) NPS, and c) APS \cite{Pushing_corridor}. The plots depict the average minimum distance, denoting the distance between the contact point where the object touches the robot and the goal location, for both sets of friction across every target position and object. The results strongly demonstrate the superior performance of our RPS especially in comparison to NPS and APS.}
	\label{fig:min_distance}
  \vspace{-0.4cm}
\end{figure*}

\subsection{The Simulation Experiments}\label{subsec:simulation_experiments}

The virtual environment was developed using the Gazebo simulator version 11.11 with the Open Dynamics Engine (ODE), where a virtual model of the Kairos platform was used to simulate the mobile manipulator. In order to simulate the capacitive touch skin, we attached a contact sensor provided by the Gazebo environment to the base of our virtual mobile robot~\cite{contactsensor}. Despite this sensor providing the wrench information, we opted to utilize only the position information of the contact, in alignment with Sec.~\ref{sec:touch_skin_sensor}.

To demonstrate the robustness of the proposed strategy, we conducted virtual experiments involving two sets of friction coefficients (see Table \ref{tab:friction}) and three objects with varying masses and shapes (see Fig. \ref{fig:objects} for their characteristics). Each combination of object-friction set was targeted to push to 24 points on the X-Y plane, spaced at 1-meter intervals from 2 to -2 for each axis, excluding (0,0) where the robot is located at the beginning of the experiments (see Fig. \ref{fig:min_distance}). Target points behind and right next to the robot were intentionally selected to assess the strategy effectiveness in these challenging scenarios. To consider an object-pushing trial as successful, the robot must push the object 0.05 meters (${d}_{success}$) close to the target within 300 seconds, without losing contact for more than 150 seconds.
\begin{table}[h]\vspace{-1mm}
\centering
\caption{The sets of friction coefficients}\vspace{-2mm}
\begin{tabular}{|M{2.0cm}|M{1.0cm}|M{1.0cm}|}
\hline
 & $\mathcal{S}_{\mu1}$& $\mathcal{S}_{\mu2}$ \\
\hline
$\mu_{object-ground}$ & 0.3 & 0.2 \\
\hline
$\mu_{object-robot}$ & 0.35 & 0.5 \\
\hline
\end{tabular}\vspace{-4mm}
\label{tab:friction}
\end{table}

The performance of our RPS during the described scenarios was assessed by comparing it with a) a non-reactive pushing strategy (NPS), and b) an existing method from the literature namely adaptive pushing skill (APS)~\cite{Pushing_corridor}. The non-reactive strategy refers to a version of our approach that excludes the lateral velocities ($v^{R}_{y}$), which are utilized to induce a sliding behavior for the object toward the center of the robot, by setting the adaptive rate ($a_{r}$) to 0 throughout the task. It's important to note that this approach also lacks the \textit{Realignment State}, which is a key feature of our proposed strategy that ensures not to lose contact with the pushed object when it is close to the edges of the robot front side. Among the mobile pushing frameworks discussed in Sec.~\ref{sec:introduction}, the adaptive pushing skill~\cite{Pushing_corridor} stands out as the suitable baseline due to its capability to push unknown objects toward desired targets without requiring prior knowledge of friction distributions and mass properties, akin to our strategy. This skill, validated with a circular-base mobile robot, learns the dynamic behavior of the pushed object on the fly by using vision-based data of the object motion.  Then, the learned model is utilized to blend actions of relocating and pushing, which generates a new angle of attack when desired.

Fig.~\ref{fig:min_distance} depicts the minimum distance achieved between the contact point where the object touches the robot and the target location throughout the scenarios for all three methods described previously. For each object, the minimum distance is presented as the average value for both friction sets at every target position. During these scenarios, if this distance reaches the 0.05 m threshold under the conditions described previously, the trial is ended and considered successful. The values at every target point indicate the number of successful trials out of 6 trials (comprising 3 different objects and 2 friction sets).

These results show that the proposed strategy, RPS, has the highest success rate, with 88.19\% of the trials meeting the success criterion. However, the baseline strategies, NPS and APS, achieved the goal of pushing the object to the target point under the 0.05m threshold only in 15.97\% and 5.56\% of the trials, respectively. The color mapping reveals that even in the unsuccessful trials under RPS, the object came very close to the target point, demonstrating the effectiveness of our approach, except for the scenarios involving the nonuniform box with the (0,1) target point. Furthermore, our proposed strategy stands out as the only method capable of pushing the object toward the target points located behind (14 locations from 0 to -2 on the X axis). It's worth mentioning that NPS was successful in some scenarios, being feasible without having sharp turns while pushing. In contrast, APS was limited to delivering objects solely towards targets directly positioned in front of the initial location as also reported in~\cite{unwieldy}, where tested with a square-based robot instead of a circular-shaped as in the original study~\cite{Pushing_corridor}.

\begin{figure*}[!t]	
\centering
	\includegraphics[width=0.9\textwidth]{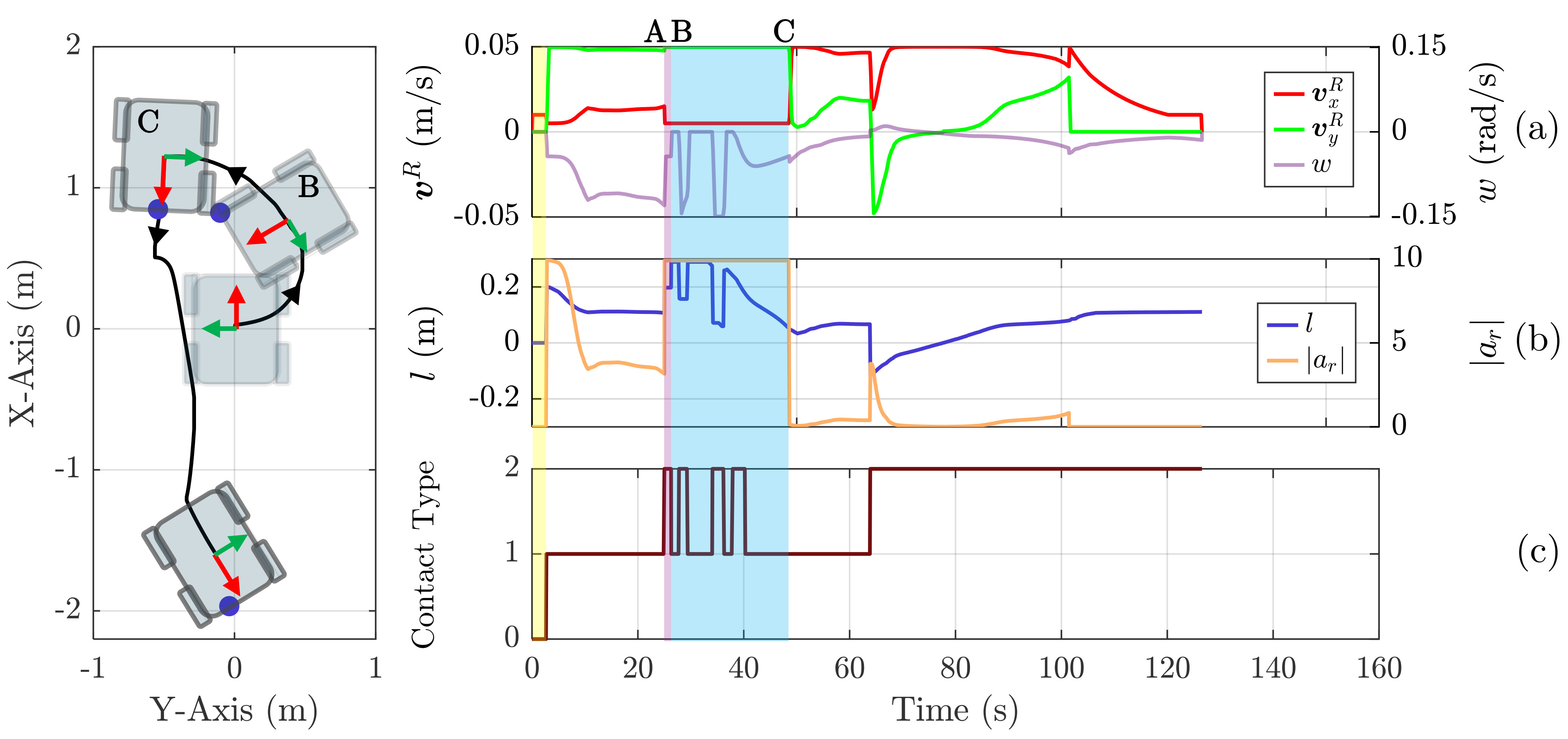}
    \vspace{-0.2cm}

	\caption{The results of an experimental scenario with the first friction set ($\mathcal{S}_{\mu1}$) where the box was pushed to the target (-2, 0) using the reactive pushing strategy (RPS). While the left column plot displays the path of the mobile platform as it navigates towards the target with the blue point indicating the contact location, the corresponding plots on the right present a) the velocity of the robot ($\boldsymbol{v}^{R}$), b) the contact position ($l$) and the adaptive rate ($a_r$), and c) contact type (0: Contact does not exist, 1: Point Contact, 2: Line Contact) during this trial.
}
	\label{fig:example_trial}
  \vspace{-0.7cm}
\end{figure*}

\begin{table}[!t]
\centering
\caption{Trial Success Rate }\vspace{-2mm}
\begin{tabular}{|M{0.5cm}|M{0.65cm}|M{0.65cm}|M{0.65cm}|M{0.65cm}|M{1cm}|M{1cm}|}
\cline{1-7}
&\multicolumn{2}{c|}{\textbf{Box}} & \multicolumn{2}{c|}{\textbf{Cylinder}} & \multicolumn{2}{c|}{\textbf{Nonuniform Box}}  \\
\hline
& $\mu_{1}$ & $\mu_{2}$ & $\mu_{1}$ & $\mu_{2}$ & $\mu_{1}$ & $\mu_{2}$  \\
\hline
RPS&23/24 & 24/24 & 19/24 & 19/24 & 19/24 & 23/24 \\
\hline
NPS &2/24 & 5/24 & 2/24 & 2/24 & 6/24 & 6/24 \\
\hline
APS &2/24 & 2/24 & 2/24 & 2/24 & 0/24 & 0/24 \\
\hline
\end{tabular}
\label{Table:success_Rate}
 \vspace{-0.5cm}
\end{table}

The success rates of the pushing strategies across different object-friction sets are reported in Table~\ref{Table:success_Rate}. Although our proposed strategy exhibited the highest success rate overall for each object-friction set, the scenarios involving the cylinder have a slightly lower rate. The success rate for the nonuniform box under the non-reactive pushing strategy was higher than for the other objects, since it could also be pushed to the front-left target points besides the targets directly in front, as seen in Fig.~\ref{fig:min_distance}. On the other hand, the APS could not achieve to push the nonuniform box in any of the targets.

Fig.~\ref{fig:example_trial} demonstrates the results of a scenario where the robot pushed the box to the target (-2, 0) with the first friction set. In this figure, the left plot illustrates the path traced by the robot odometry, with the blue dot indicating the point where the box made contact with the capacitive touch sensor. It is important to note that, as explained previously, this point was calculated by averaging the corner locations when line contact occurs. The corresponding plots on the right display the results collected from this trial.

At the beginning of the experiment reported in Fig. 6, the robot moved forward at a relatively low velocity ($v_x^R = 0.01$ m/s) until it made contact with the object (yellow highlighted region). Despite the box being placed in front of the robot without any rotation, the contact sensor initially detected only the right corner of the object, as seen from the contact type and location (Fig.~\ref{fig:example_trial}b and Fig.~\ref{fig:example_trial}c). Since the detected corner was located 0.2 m from the center and near the border of the robot front edge, the adaptive rate ($a_r$) started from nearly at its highest value, leading to high lateral velocities. During this motion, the robot also rotated with a relatively high angular velocity to turn the object towards the target point, which was initially located behind the robot. Although the initial contact point started to slide to the center of the robot front side, as intended, a line contact occurred with the side edge of the box due to the robot rotation around the object (A, the beginning of the pink highlighted region). This shift in the contact type from a point to a line resulted in a sudden change in the contact location, causing the adaptive rate to increase its highest value again. Shortly after this, due to the ongoing angular motion of the base, the contact point, which was closer to the robot center, lost contact. This left only the point near the edge of the robot in contact with the object. Since this point exceeds the threshold that specifies the critical regions at the edges, the robot activated the \textit{Realignment State} (B, the beginning of the blue highlighted region). During the blue highlighted region, this state remained active until it completed re-positioning the contact point to the center of the robot (C, the end of the blue highlighted region), where the angular velocities were calculated using Eq.~\ref{eq:w_rfm}. Finally, the robot successfully pushed the object to the target along with lateral movements, and these reactive motions prevented the object from entering any of the critical regions throughout the remaining task.

\subsection{The Real-world Experiments}\label{subsec:hardware_experiments}

The real-world experiments have been conducted using the proposed RPS with two different unwieldy objects: a) a 25 kg Box with dimensions of 32 $\times$ 26 $\times$ 47 cm, and b) a 12 kg cylinder with a radius of 11.5 cm and a height of 49 cm. The video displaying the pushing of these objects, including one with an additional bulky closet and another demonstration where the object was displaced intentionally during pushing, is available at \url{https://youtu.be/IuGxlNe246M}. Each object was pushed to 8 different target points at 2.1 m intervals apart from the initial position of the robot (0,0). 

\begin{figure}[t]
\centering
	\includegraphics[width=0.8\columnwidth]{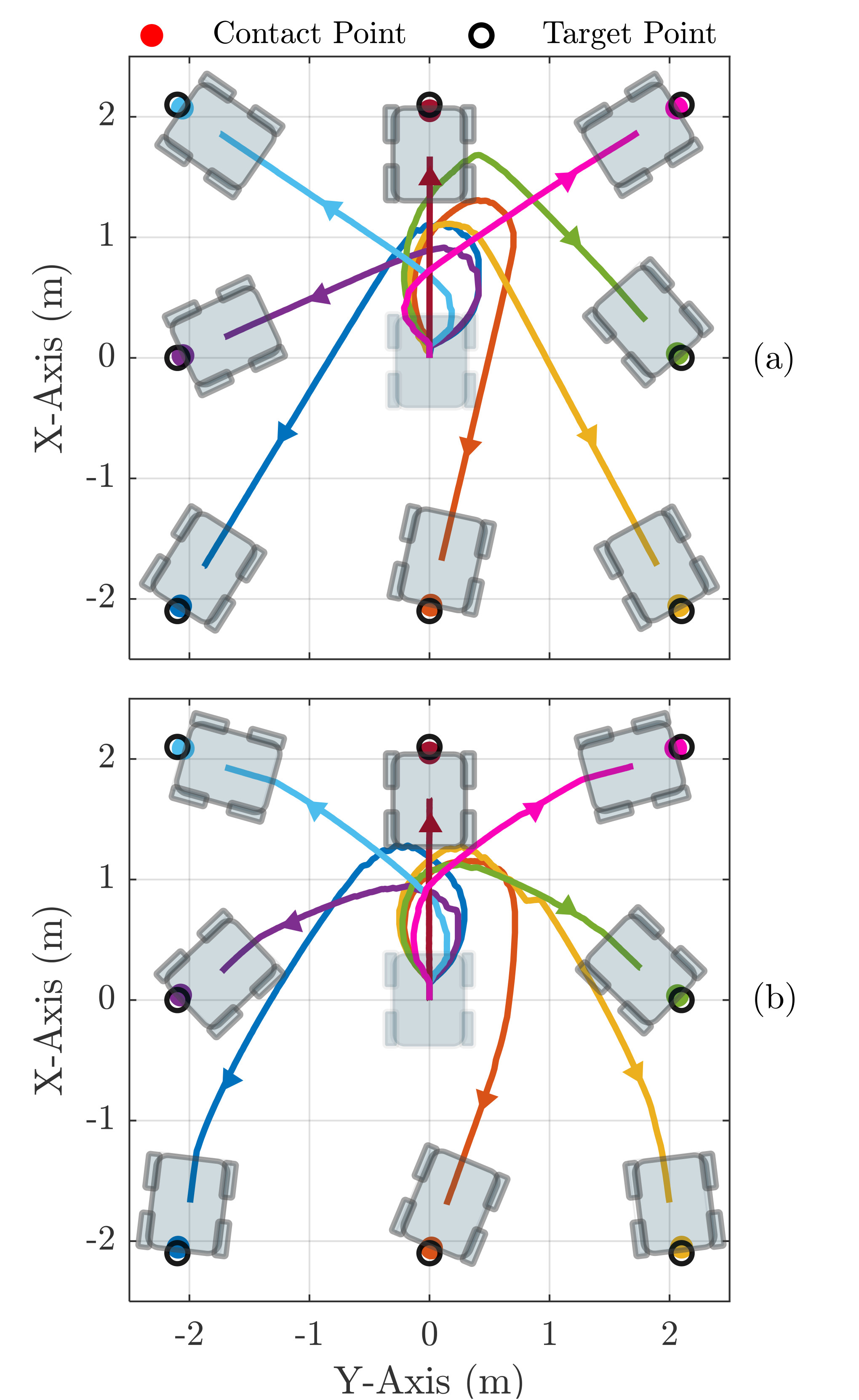}
	\caption{The paths followed by the mobile base in the real-world experiments carried out with a) the box and b) the cylinder using the RPS.}
	\label{fig:harware_exp}
 \vspace{-0.7cm}
\end{figure}

Fig.~\ref{fig:harware_exp} presents the paths of the mobile base, the target locations, and the contact points at the end of the experiment. Our proposed strategy achieved the objective of pushing the object to targets close to 0.05 m with a 100\% success rate, including the challenging ones behind and to the sides, without prior knowledge of the mass and frictional parameters related to the object and environment. In these experiments, the average of all scenarios for the mean deviation of the contact point from the center ($D$) throughout the task was obtained as 0.032 and 0.059 m for the box and the cylinder, respectively. This deviation was calculated for each scenario as follows:
\begin{equation}
     D = 1/({t_{e}-t_{s}})\int_{t_{s}}^{t_{e}}{|l|dt},
\end{equation}
where $t_s$ and $t_e$ indicate the starting and ending time of the experiment, and $|l|$ denotes the absolute value of the lateral distance between the contact location and the robot center. Additionally, the average completion time of the scenarios closely approximated each other, with durations of 89 and 88 seconds for the box and the cylinder, respectively.

\section{Discussion and Conclusion}
\label{sec:discussion_and_conclusion}

In this paper, we introduced a novel pushing strategy capable of generating reactive robot movements that effectively drive the object toward a goal location, neither relying on prior assumptions of object-specific properties nor utilizing modeled object behavior during pushing. 
To systematically assess the robustness of the \textit{Reactive Pushing Strategy}, experiments were conducted in both virtual and real-world environments. According to the simulation results, the proposed strategy exhibited a success rate almost five times higher than that of its closest competitor, the non-reactive one (NPS).
Upon further analysis of Fig. \ref{fig:min_distance}, it can be seen that the RPS significantly outperforms both the NPS and APS in terms of the distance remaining to reach the goal point in scenarios where the target was not reached.

Moreover, the results obtained for the RPS involving various object-friction sets (refer to Table \ref{tab:friction}) reveal that the scenarios involving the cylinder exhibited a slightly lower success rate than the remaining objects; nonetheless, our strategy consistently outperformed the baselines.
This lower success rate may stem from two main factors: the rolling behavior of the cylinder and the establishment of line contact with the nonuniform and uniform boxes, contributing to a more stable push manipulation \cite{Lynch_Mason}.
Here, we would like to emphasize that a notable strength of the \textit{Reactive Pushing Strategy} is its capability to eliminate the flat bumper problem introduced by Igarashi et al. \cite{Igarashi2010}. 
The authors concluded that rotational motion tends to slide the object in an undesired direction for robots with flat bumpers.
To address this challenge, the robot holonomic motion capability is leveraged to adjust angular and lateral velocities dynamically.
In contrast to the studies that employ circular-shaped robots to avoid this issue, we showed that our proposed system can possess both point and line contact during object manipulation and is robust to the transitions between these contact types (see the shaded blue area in Fig. \ref{fig:example_trial}c).

Furthermore, the real-world experiments validated that, even without knowledge of object and environment-related information, our push manipulation method successfully reached all target points, including challenging ones positioned behind and to the sides. In a recent study \cite{unwieldy} employing rectangular-shaped mobile robots akin to ours, successful object deliveries were achieved only for targets initially positioned in front. The limited range of target points in their work could be anticipated due to the use of nonholonomic robots and the imposition of a stiff line contact constraint to avoid repositioning actions, which can be time-consuming with the nonholonomic motion constraint.

Overall, our main objective while formulating the \textit{Reactive Pushing Strategy} is to minimize reliance on additional information, facilitating its application in unstructured and dynamic environments.
Approaches that depend on prior knowledge have generally been constrained to laboratory settings, as the assumptions upon which they are built need highly structured environments to be held. 
On the other hand, modeling or learning-based approaches typically rely on vision-based measurements, where the accuracy in tracking objects and the occurrence of occlusion issues limit the performance \cite{LetsPushReview, Pushing_corridor}.

\vspace{-0.3cm}

\bibliographystyle{IEEEtran}
\bibliography{biblio}

\end{document}